\documentclass[10pt,twocolumn,letterpaper]{article}

\usepackage{iccv}
\usepackage{times}
\usepackage{epsfig}
\usepackage{graphicx}
\usepackage{amsmath}
\usepackage{amssymb}
\usepackage{float}
\usepackage{booktabs,multirow}
\usepackage{enumitem}
\usepackage{lipsum}
\usepackage[T1]{fontenc}
\usepackage[accsupp]{axessibility}

\usepackage[pagebackref=true,breaklinks=true,letterpaper=true,colorlinks,bookmarks=false]{hyperref}
\setlist[itemize]{noitemsep, topsep=0pt}

\iccvfinalcopy 

\def\iccvPaperID{10829} 

\ificcvfinal\fi

\begin{document}

\title{SVDFormer: Complementing Point Cloud via Self-view Augmentation and Self-structure Dual-generator}

\author{Zhe Zhu$^{1}$, Honghua Chen$^{1}$, Xing He$^1$, Weiming Wang$^{2\dag}$, Jing Qin$^3$, Mingqiang Wei$^{1\dag}$\\
$^1$Nanjing University of Aeronautics and Astronautics\\
$^2$Hong Kong Metropolitan University\\
$^3$The Hong Kong Polytechnic University\\
{\tt \small zhuzhe0619@nuaa.edu.com;} {\tt \small chenhonghuacn@gmail.com;} {\tt \small hexing@nuaa.edu.cn;}\\
{\tt \small wmwang@hkmu.edu.hk;} {\tt \small harry.qin@polyu.edu.hk;} 
{\tt \small mqwei@nuaa.edu.cn}
}
\twocolumn[{%
\maketitle
\begin{figure}[H]
    \hsize=\textwidth
    \centering
    \includegraphics[width=\textwidth]{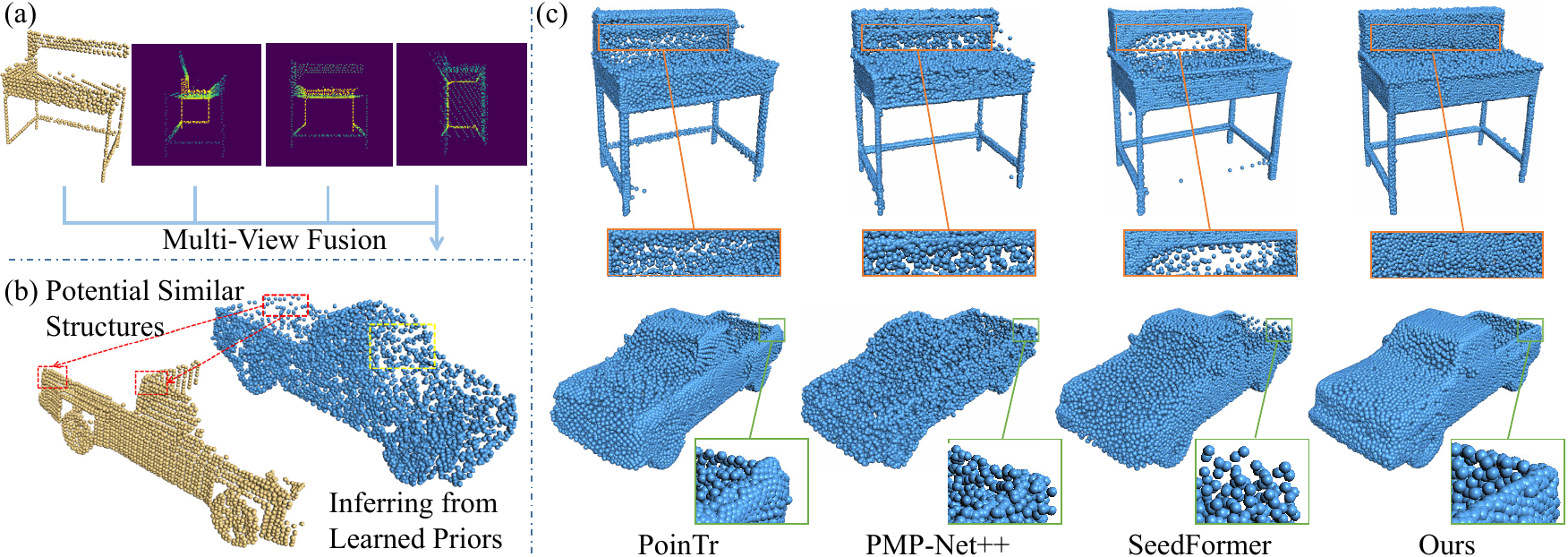}
    \caption{The characteristics of our method and visual comparison of point cloud completion results. (a) SVDFormer understands incomplete shapes from self-projected multiple views. (b) SVDFormer collaborates both geometric similarities (red boxes) and shape priors (yellow boxes) for shape refinement. (c) Qualitative comparison of our SVDFormer with PoinTr~\cite{yu2021pointr}, PMP-Net++~\cite{wen2022pmp}, and SeedFormer~\cite{zhou2022seedformer}.}
    \label{fig:teaser}
\end{figure}
}]


\ificcvfinal\thispagestyle{empty}\fi

\renewcommand{\thefootnote}{}
\footnotetext{ \dag Co-corresponding authors}
\begin{abstract}
In this paper, we propose a novel network, SVDFormer, to tackle two specific challenges in point cloud completion: understanding faithful global shapes from incomplete point clouds and generating high-accuracy local structures. Current methods either perceive shape patterns using only 3D coordinates or import extra images with well-calibrated intrinsic parameters to guide the geometry estimation of the missing parts. However, these approaches do not always fully leverage the cross-modal self-structures available for accurate and high-quality point cloud completion.
To this end, we first design a Self-view Fusion Network that leverages multiple-view depth image information to observe incomplete self-shape and generate a compact global shape.
To reveal highly detailed structures, we then introduce a refinement module, called Self-structure Dual-generator, in which we incorporate learned shape priors and geometric self-similarities for producing new points.
By perceiving the incompleteness of each point, the dual-path design disentangles refinement strategies conditioned on the structural type of each point.
SVDFormer absorbs the wisdom of self-structures, avoiding any additional paired information such as color images with precisely calibrated camera intrinsic parameters.
Comprehensive experiments indicate that our method achieves state-of-the-art performance on widely-used benchmarks.
Code is available at \url{https://github.com/czvvd/SVDFormer}.
\end{abstract}

\section{Introduction}
Point cloud completion plays an essential role in 3D vision applications and remains an active research topic in recent years. 
To tackle this task, a variety of learning-based techniques have been proposed, of which many demonstrate encouraging results~\cite{yuan2018pcn,huang2020pf,zhang2020detail,yu2021pointr,xiang2021snowflakenet,yan2022fbnet,zhang2022shape,tang2022lake,zhou2022seedformer,zhang2022point}. 
However, the sparsity and large structural incompleteness of captured point clouds still limit the ability of these methods to produce satisfactory results.

We observe that there are two primary challenges in this task.
The first challenge is that crucial semantic parts may be absent, resulting in a vast solution space for point-based networks~\cite{yuan2018pcn,xiang2021snowflakenet,yu2021pointr,zhou2022seedformer} to identify plausible global shapes and locate missing regions.
Some alternative methods attempt to address this issue by incorporating additional color images~\cite{zhang2021view,aiello2022crossmodal,zhu2023csdn}, but the paired images are hard to obtain, as well as the well-calibrated intrinsic parameters.
The second one is how to infer detailed structures. 
Some recent methods~\cite{xiang2021snowflakenet,yan2022fbnet} utilize skip-connections between multiple refinement steps, allowing them to better leverage learned shape pattern priors to iteratively recover finer details. 
Some other methods prioritize preserving the original detail information by no-pooling encoding~\cite{zhou2022seedformer} or structural relational enhancement~\cite{pan2021variational}.
However, all above mentioned approaches typically employ a unified refinement strategy for all surface regions, which hinders the generation of geometric details for various missing regions. 
By observing and analyzing the partial inputs, we find that the missing surface regions can be classified into two types.
The first type lacks similar structures in the input shape, and their reconstruction heavily relies on the learned shape prior.
The second type is consistent with the local structures that are present in the partial input, and their recovery can be facilitated by appropriate geometric regularity~\cite{zhao2021sign}. 
For instance, LiDAR scans in KITTI~\cite{geiger2013vision} are highly sparse and contain limited information for generating fine details. Existing refinement strategies tend to produce and preserve implausible line-like shapes (see from Figure~\ref{fig:real}). 

Based on the above observations, we propose a new neural network for point cloud completion called SVDFormer.
Our method makes improvements by fully leveraging self-structure information in a coarse-to-fine paradigm. 

First, similar to how a human would perceive and locate the missing areas of a physical object by observing it from different viewpoints, we aim to drive the neural network to absorb this knowledge by augmenting the data representation. To achieve this, we design a Self-View Fusion Network (SVFNet) that learns an effective descriptor, well depicting the global shape from both the point cloud data and depth maps captured from multiple viewpoints (see from Figure~\ref{fig:teaser} (a)). To better exploit such kind of cross-modal information, we specifically introduce a feature fusion module to enhance the inter-view relations and improve the discriminative power of multi-view features.

Regarding the second challenge, our insight is to disentangle refinement strategies conditioned on the structural type of each point to reveal detailed geometric structures.
Therefore, we design a Self-structure Dual-Generator (SDG) with a pair of parallel refinement units, called \emph{Structure Analysis} and \emph{Similarity Alignment}, respectively. 
The former unit analyzes the generated coarse point clouds by explicitly encoding local incompleteness, which enables it to match learned geometric patterns of training data to infer underlying shapes.
The \emph{Similarity Alignment} unit finds the features of similar structures for every point, thus making it easier to refine its local region by mimicking the geometry of input local structures. 
With the aid of this dual-path design, our method can generate reasonable results for different types of input shapes, including symmetrical synthetic models with various degrees of incompleteness and highly sparse real-world scans.

Extensive experiments demonstrate that SVDFormer achieves state-of-the-art performance on widely-used benchmarks. Our key contributions are listed below:
\begin{itemize}
  \item We design a novel network called SVDFormer, which significantly improves point cloud completion in terms of global shape understanding and details recovery.
  \item We propose the novel Self-view Fusion Network (SVFNet) equipped with a feature fusion module to enhance the multi-view and cross-modal feature, 
  which can output a plausible global shape.
  \item We introduce a Self-structure Dual-Generator (SDG) for refining the coarse completion. It enables our method to handle various kinds of incomplete shapes by jointly learning the local pattern priors and self-similarities of 3D shapes.
\end{itemize}
\label{secIntroduction}

\section{Related Work}
\begin{figure*}[ht]
  \centering
  \includegraphics[width=\textwidth]{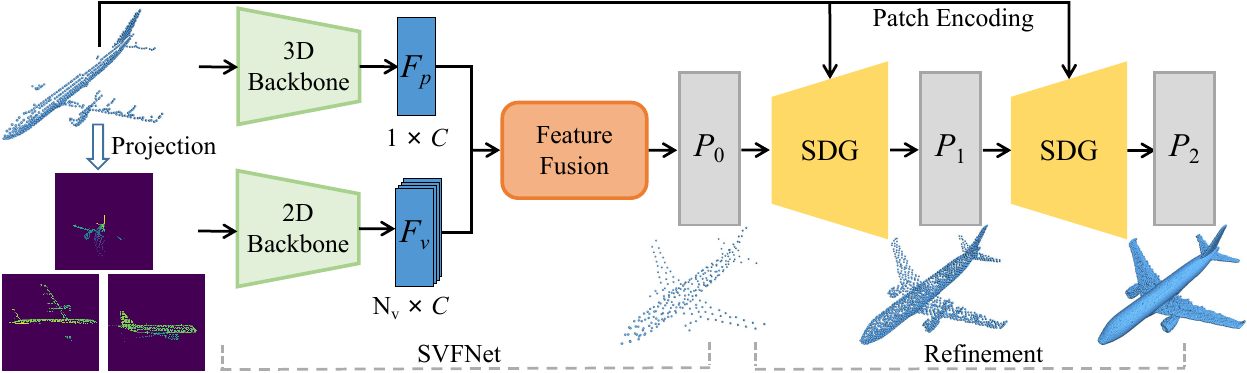}
\caption{The architecture of SVDFormer. SVFNet first generates a global shape from the cross-modal input. The coarse completion is then upsampled and refined with two SDGs.}
  \label{fig:overview}
\end{figure*}
\subsection{Learning-based Shape Completion}
Early learning-based methods ~\cite{dai2017shape,han2017high,stutz2018learning,varley2017shape} often rely on voxel-based representations for 3D convolutional neural networks. However, these approaches are limited by their high computational cost and limited resolution. Alternatively, GRNet~\cite{xie2020grnet} and VE-PCN~\cite{wang2021voxel} use 3D grids as an intermediate representation for point-based completion. 

In recent years, several methods are proposed to directly process points by end-to-end networks. One pioneering point-based work is PCN~\cite{yuan2018pcn}, which uses a shared multi-layer perceptron (MLP) to extract features and generates additional points using a folding operation~\cite{yang2018foldingnet} in a coarse-to-fine manner. Inspired by it, a lot of point-based methods~\cite{wang2020cascaded,liu2020morphing,wen2020point,xiang2021snowflakenet,zhou2022seedformer,yu2021pointr} have been proposed.

Later, to address the issue of limited information available in partial shapes using only point data, several works~\cite{zhang2021view,aiello2022crossmodal,zhu2023csdn,huang2022spovt} have explored the use of auxiliary input to enhance performance. We call them cross-modal-based methods. These approaches involve the combination of rendered color images and partial point clouds, along with the corresponding camera parameters.
Although these methods have shown promising results, they often require additional input that is difficult to obtain in practical settings.
Different from these 3D data-driven methods, MVCN~\cite{hu2019render4completion} operates completion solely in the 2D domain using a conditional GAN. However, it lacks the ability to supervise the results using ground truth with rich space information.
Besides, some other methods~\cite{xie2021style,aiello2022crossmodal} seek to supervise point cloud completion in the 2D domain. The 2D projections of completed points are used to calculate loss by comparing them to the ground-truth depth. 
In contrast to these methods, we propose to utilize 2D input by observing self-structures to understand the overall shape. As a result, our method achieves a more comprehensive perception of the overall shape without requiring additional information or differentiable rendering during training.

Considering the high-quality details generation, a variety of strategies have been introduced by learning shape context and local spatial
relationships. 
To achieve this goal, state-of-the-art methods design various refinement modules to learn better shape priors from the training data.
SnowflakeNet~\cite{xiang2021snowflakenet} introduces Snowflake Point Deconvolution (SPD), which leverages skip-transformer to model the relation between parent points and child points. 
FB-Net~\cite{yan2022fbnet} adopts the feedback mechanism during refinement and generates points in a recurrent manner.
LAKe-Net~\cite{tang2022lake} integrates its surface-skeleton representation into the refinement stage, which makes it easier to learn the missing topology part.
Another type of method tends to preserve and exploit the local information in partial input.
One direct approach is to predict the missing points by combining the results with partial input data~\cite{huang2020pf,yu2021pointr}. As the point set can be viewed as a token sequence, PoinTr~\cite{yu2021pointr} employs the transformer architecture~\cite{vaswani2017attention} to predict the missing point proxies.
SeedFormer~\cite{zhou2022seedformer} introduces a shape representation called patch seeds for preventing the loss of local information during pooling operation. 
Some other approaches~\cite{pan2020ecg,pan2021variational,zhang2022point} propose to enhance the generated shapes by exploiting the structural relations in the refinement stage.
However, these strategies employ a unified refinement strategy for all points, which limits their ability to generate pleasing geometric details for different points.
Our approach differs from theirs by breaking down the shape refinement task into two sub-goals, and adaptively extracting reliable features for different partial areas.

\subsection{Multi-view fusion for Shape Learning}
View-based 3D shape recognition techniques have gained significant attention in recent years.
The classic Multi-View Convolutional Neural Network (MVCNN) model was introduced in \cite{su2015multi}, where color images are fed into a CNN and subsequently combined by a pooling operation. However, this approach has the fundamental drawback of ignoring view relations.
Following works~\cite{feng2018gvcnn,wei2022learning,han20193d2seqviews,yang2019learning} propose various strategies to tackle this problem. For example,
Yang \emph{et al.}~\cite{yang2019learning} obtains a discriminative 3D object representation by modeling region-to-region relations.
LSTM is used to build the inter-view relations~\cite{dai2018siamese}.
Since the cross-modal data are more available recently, methods are proposed to fuse features of views and point clouds~\cite{you2018pvnet,you2019pvrnet}.
Inspired by the success of multi-view fusion, our method utilizes point cloud features to enhance relationships between multiple views obtained by self-view augmentation.

\label{secRelatedWork}
\section{Method}
The input of our SVDFormer consists of three parts: a partial and low-res point cloud $P_{in}\subseteq\mathbb{R}^{N\times3}$, $N_V$ camera locations $VP\subseteq\mathbb{R}^{N_V\times3}$ (three orthogonal views in our experiments), and $N_V$ depth maps $D\subseteq\mathbb{R}^{N_V\times1\times H \times W}$.
Given these inputs, our goal is to estimate a complete point cloud $P_{2}\subseteq\mathbb{R}^{N_2\times3}$ in a coarse-to-fine manner.
The overall architecture is exhibited in Figure~\ref{fig:overview}, which comprises two parts: an SVFNet and a refiner equipped with two SDG modules. 
SVFNet first leverages multiple self-projected depth maps to produce a globally completed shape $P_0\subseteq\mathbb{R}^{N_0\times3}$. Subsequently, two SDGs gradually refine and upsample $P_0$ to yield the final point cloud $P_{2}$, which exhibits geometric structures with high levels of detail.
Note that unlike some recent cross-modal approaches~\cite{zhang2021view,zhu2023csdn,aiello2022crossmodal,zhang2022shape}, \textit{our method makes full use of self-structures and does not require any additional paired information such as color images with precisely calibrated camera intrinsic parameters~\cite{zhang2021view,zhu2023csdn}}. Depth maps are directly yielded by projecting point clouds themself from controllable viewpoints during the data-preprocessing stage.

\subsection{SVFNet}
The SVFNet aims to observe the partial input from different viewpoints and learns an effective descriptor to produce a globally plausible and complete shape. We first extract a global feature $F_p$ from $P_{in}$ using a point-based 3D backbone network and a set of view features $F_V$ from the $N_V$ depth maps using a CNN-based 2D backbone network. We directly adopt well-established backbone networks. In detail, the PointNet++~\cite{qi2017pointnet++} with three set abstraction layers encodes $P_{in}$ in a hierarchical manner and the ResNet-18 model~\cite{he2016deep} is employed as the 2D backbone.

However, how to effectively fuse the above cross-modal features is challenging.
In our early experiments, we directly concatenate these features, but the produced shape is less pleasing (see the ablation studies in Section~\ref{ablationSec}). This may be caused by the domain gap between 2D and 3D representations.
To resolve this problem, we propose a new feature fusion module, to fuse $F_p$ and $F_V$, and output a global shape descriptor $F_g$, followed by a decoder to generate the global shape $P_c$. The decoder uses a 1D Conv-Transpose layer to transform $F_g$ to a set of point-wise features and regresses 3D coordinates with a self-attention layer~\cite{vaswani2017attention}. Finally, we adopt a similar approach to previous studies~\cite{xiang2021snowflakenet,zhou2022seedformer}, where we merge $P_c$ and $P_{in}$ and resample the merged output to generate the coarse result $P_0$.
\begin{figure}[t]
  \centering
  \includegraphics[width=\linewidth]{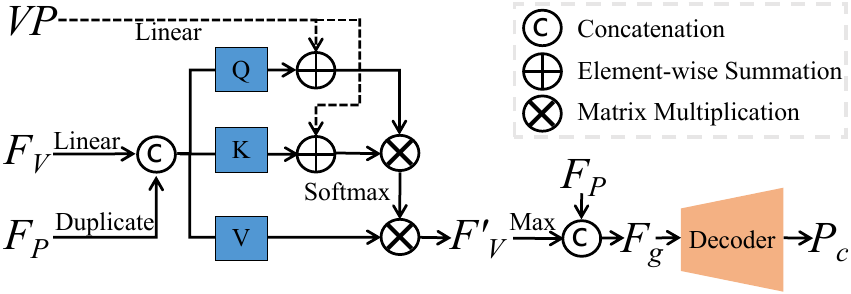}
\caption{Illustration of the feature fusion module.}
  \label{fig:VA}
\end{figure}

\noindent\textbf{Feature Fusion.}
As shown in Figure~\ref{fig:VA}, $F_V$ is first transformed to query, key, and value tokens via linear projection and the guidance of global shape feature $F_P$. 
Then, to enhance the discriminability of view features, the attention weights are calculated based on the query and key tokens conditioned on the projected viewpoints $VP$. Detailedly, we map $VP$ into the latent space through a linear transformation and then use them as positional signals for feature fusion.
After the elemental-wise product, each feature in $F_V'$ combines the relational information from other views under the guidance of $F_P$. Finally, the output shape descriptor $F_g$ is derived from $F_V'$ via maximum pooling.
\begin{figure*}[t]
  \centering
  \includegraphics[width=\textwidth]{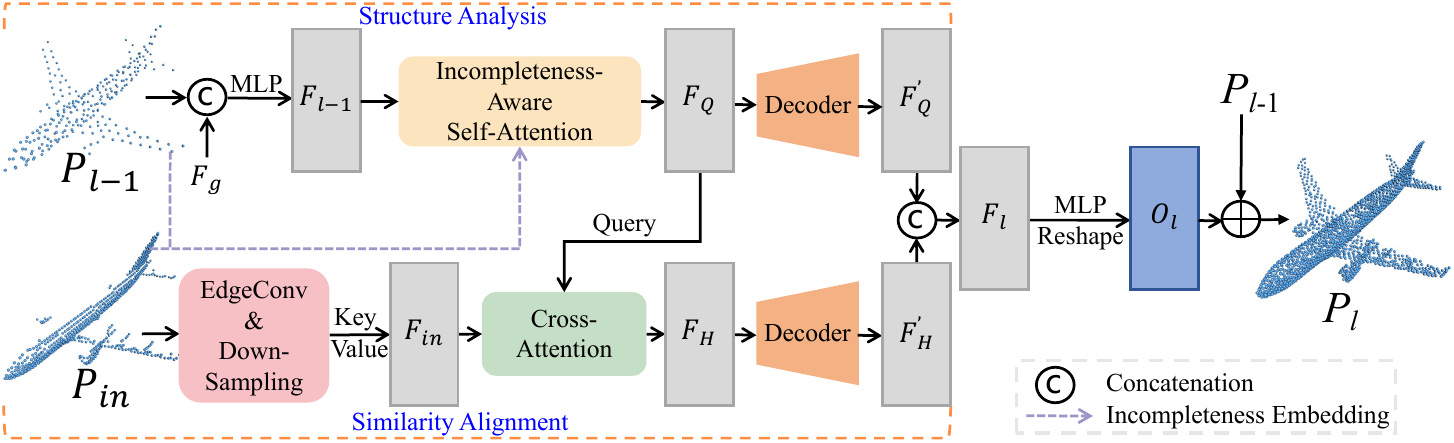}
\caption{The architecture of SDG. The upper path represents Structure Analysis and the lower path represents Similarity Alignment. Each sub-network generates an offset feature which is then concatenated with each other and used to regress into the coordinate offsets.}
  \label{fig:IDTr}
\end{figure*}
\subsection{SDG}
The SDG seeks to generate a set of coordinate offsets to fine-tune and upsample the coarse shape, based on the structural type of the missing surface region. 
To achieve it, SDG is designed as a dual-path architecture as shown in Figure~\ref{fig:IDTr}, which consists of two parallel units named \emph{Structure Analysis} and \emph{Similarity Alignment}, respectively. Overall, fed with the partial input $P_{in}$ and coarse point cloud $P_{l-1}$ outputted in the last step, we obtain the combined point-wise feature $F_{l}\subseteq\mathbb{R}^{N\times 2C}$. $F_{l}$ comprises two kinds of sources of shape information: one is derived from learned shape priors, while the other is learned from the similar geometric patterns found within $P_{in}$. $F_{l}$ is then projected to a higher dimensional space and reshaped to produce a set of up-sampled offsets $O_{l}\subseteq\mathbb{R}^{rN\times 3}$, where $r$ represents the upsampling rate. 
The predicted offsets are further added back to $P_{l-1}$ to obtain a new completion result. Note that we iterate SDG twice, as shown in Figure~\ref{fig:overview}.

\subsubsection{Structure Analysis}
Since detailed geometries from missing regions are harder to be recovered, we embed an incompleteness-aware self-attention layer to explicitly encourage the network to focus more on the miss regions. Specifically, $P_{l-1}$ is first concatenated with the shape descriptor $F_g$, and then embedded into a set of point-wise feature $F_{l-1}=\{f_i\}^{N_{l-1}}_{i=1}$ by a linear layer. 
Next, $F_{l-1}$ is fed to the incompleteness-aware self-attention layer to obtain a set of features $F_Q=\{q_i\}^{N_{l-1}}_{i=1}$, which encodes the point-wise incompleteness information. 
$q_i$ is computed by:
\begin{equation} \label{eqn1}
  \begin{split}
   q_i &= \sum_{j=1}^{N_{l-1}}{a_{i,j}(f_jW_V)}\\
   a_{i,j} &= Softmax((f_iW_Q+h_i)(f_jW_K+h_j)^T)
  \end{split},
\end{equation}
where $W_Q$, $W_K$, and $W_V$ are learnable matrix with the size of $C \times C$. $h_i$ is a vector that represents the degree of incompleteness for each point $x$ in $P_{l-1}$. 
Intuitively, points in missing regions tend to have a larger distance value to the partial input. We thus calculate the incompleteness by:
\begin{equation} \label{eqn2}
  \begin{split}
   h_i &= Sinusoidal(\frac{1}{\gamma}\min_{y \in P_{in}}\lvert \lvert x - y \lvert\lvert)
  \end{split},
\end{equation}
where $\gamma$ is a scaling coefficient. We set it as 0.2 in our experiment.
The sinusoidal function~\cite{vaswani2017attention} is used to ensure that $h_i$ has the same dimension as the embeddings of query, key, and value. We then decode $F_Q$ into $F_{Q}^{'}$ for further analysis of the coarse shape.

\subsubsection{Similarity Alignment}
The \emph{Similarity Alignment} unit exploits the potential similar local pattern in $P_{in}$ for each point in $P_{l-1}$ and addresses the feature mismatch problem caused by the unordered nature of point clouds.
Inspired by the point proxies in \cite{yu2021pointr}, we begin by using three EdgeConv layers~\cite{wang2019dynamic} to extract a set of downsampled point-wise feature $F_{in}$. Each vector in $F_{in}$ captures the local context information.
Since there could exist long-range similar structures, we then perform feature exchange by cross-attention, which is a classical solution for feature alignment.
The calculation process is similar to vanilla self-attention. The only difference lies in that the query matrix is produced by $F_Q$, while $F_{in}$ serves as the key and value vectors.
The cross-attention layer outputs point-wise feature $F_H\subseteq\mathbb{R}^{N_{l-1}\times C}$, which integrates similar local structures in $P_{in}$ for each point in the coarse shape $P_{l-1}$. 
In this way, this unit can model the geometric similarity between two point clouds and facilitate the refinement of points with similar structures in the input. 
Similar with the structure analysis unit, $F_H$ is also decoded into a new feature $F_{H}^{'}$. These two decoders have the same architecture, which is implemented with two layers of self-attention \cite{vaswani2017attention}.
For more details of the used self-attention, cross-attention, and decoders, please refer to the supplemental files.

\subsection{Loss Function}
In order to measure the differences between the generated point cloud and the ground truth $P_{gt}$, we use the Chamfer Distance (CD) as our loss function, which is a common choice in recent works. 
To facilitate the coarse-to-fine generation process, we regularize the training by computing the loss function as:
\begin{equation}
    \mathcal{L}= \mathcal{L}_{CD}(P_c,P_{gt}) + \sum_{i = {1,2}}\mathcal{L}_{CD}(P_i,P_{gt})
\end{equation}
It is worth noting that we downsample the $P_{gt}$ to the same density as $P_c$, $P_1$, $P_2$ in order to compute the losses.

\label{secOverview}

\section{Experiment}

\begin{table*}[t]
    \tiny
    \renewcommand\arraystretch{1.2}
    \centering
    \caption{Quantitative results on the PCN dataset. ({$\displaystyle \ell ^{1}$} CD $\times 10^3$ and F1-Score@1\%)}
    \label{tab:pcnl1}
    \normalsize
    \begin{tabular}{c|cccccccc|ccc}
    \toprule[1pt]
    Methods & Plane & Cabinet & Car & Chair & Lamp & Couch & Table & Boat & CD-Avg$\downarrow$  & DCD$\downarrow$  &F1$\uparrow$  \cr
    \midrule[0.3pt]
              PCN~\cite{yuan2018pcn} & 5.50 & 22.70 & 10.63 & 8.70 & 11.00 & 11.34 & 11.68 & 8.59 & 9.64& - &0.695\\
              GRNet~\cite{xie2020grnet}  & 6.45 & 10.37 & 9.45 & 9.41 & 7.96 & 10.51 & 8.44 & 8.04& 8.83& 0.622 &0.708\\
              CRN~\cite{wang2020cascaded}    & 4.79 & 9.97 & 8.31 & 9.49 & 8.94 & 10.69 & 7.81 & 8.05& 8.51& - & 0.652 \\
              NSFA~\cite{zhang2020detail}    & 4.76 & 10.18 & 8.63 & 8.53 & 7.03 & 10.53 & 7.35 & 7.48 & 8.06& - & 0.734 \\
              PoinTr~\cite{yu2021pointr}  & 4.75 & 10.47 & 8.68 & 9.39 & 7.75 & 10.93 & 7.78 & 7.29& 8.38& 0.611 & 0.745 \\
              SnowflakeNet~\cite{xiang2021snowflakenet} & 4.29 & 9.16 & 8.08 & 7.89 & 6.07 & 9.23 & 6.55 & 6.40& 7.21& 0.585 & 0.801 \\
              SDT~\cite{zhang2022point} & 4.60 & 10.05 & 8.16 & 9.15 & 8.12 & 10.65 & 7.64 & 7.66 & 8.24 & - &0.754 \\
              PMP-Net++~\cite{wen2022pmp} & 4.39 & 9.96 & 8.53 & 8.09 & 6.06 & 9.82 & 7.17 & 6.52& 7.56 & 0.611 & 0.781\\
              FBNet~\cite{yan2022fbnet}  & 3.99 & 9.05 & 7.90 & 7.38 & 5.82 & 8.85 & 6.35 & 6.18& 6.94& - & -\\
              Seedformer~\cite{zhou2022seedformer}  & 3.85 & 9.05 & 8.06 & 7.06 & \textbf{5.21} & 8.85 & 6.05 & 5.85 & 6.74 & 0.583 & 0.818 \\
              \midrule[0.3pt]
              Ours & \textbf{3.62} & \textbf{8.79} & \textbf{7.46} & \textbf{6.91} & 5.33 & \textbf{8.49} & \textbf{5.90} & \textbf{5.83}& \textbf{6.54} & \textbf{0.536} &\textbf{0.841} \\
              \bottomrule[1pt]
    \end{tabular}
\end{table*}

\begin{table} [t]
    \renewcommand\arraystretch{1.2}
    \centering
    \caption{Quantitative results on ShapeNet-55. CD-S, CD-M, and CD-H stand for CD values under the easy, moderate, and hard difficulty levels, respectively. ({$\displaystyle \ell ^{2}$} CD $\times 10^3$ and F1-Score@1\%)}
    \footnotesize
    \label{tab:shapenet55}
    \begin{tabular}{c|ccc|ccc}
    \toprule[1pt]
    \makebox[0.01\linewidth][c]{Methods} & \makebox[0.01\linewidth][c]{CD-S} & \makebox[0.01\linewidth][c]{CD-M} & \makebox[0.01\linewidth][c]{CD-H} & \makebox[0.1\linewidth][c]{CD-Avg$\downarrow$} & \makebox[0.1\linewidth][c]{DCD-Avg$\downarrow$} & \makebox[0.08\linewidth][c]{F1$\uparrow$} \cr
    \midrule[0.3pt]
              FoldingNet~\cite{yang2018foldingnet} & 2.67 & 2.66 & 4.05 & 3.12 &-&  0.082\\
              PCN~\cite{yuan2018pcn} & 1.94 & 1.96 & 4.08 & 2.66 & 0.618  & 0.133\\
              TopNet~\cite{tchapmi2019topnet} & 2.26 & 2.16 & 4.3 & 2.91 &-&0.126 \\
              PFNet~\cite{huang2020pf} & 3.83 & 3.87 & 7.97 & 5.22 &-&0.339 \\
              GRNet~\cite{xie2020grnet}  & 1.35 & 1.71 & 2.85 & 1.97 &0.592&0.238 \\
              PoinTr~\cite{yu2021pointr}  & 0.58 & 0.88 & 1.79 & 1.09 &0.575& 0.464 \\
              SeedFormer~\cite{zhou2022seedformer} & 0.50 & 0.77 & 1.49 & 0.92 &0.558& \textbf{0.472} \\
              \midrule[0.3pt]
              Ours &  \textbf{0.48} & \textbf{0.70} & \textbf{1.30} & \textbf{0.83} & \textbf{0.541} & 0.451 \\
              \bottomrule[1pt]
    \end{tabular}
\end{table}

\begin{table*}[htb]
    \renewcommand\arraystretch{1.1}
    \centering
    \caption{Quantitative results on ShapeNet-34.}
    \footnotesize
    \label{tab:shapenet34}
    \begin{tabular}{c|cccccc|cccccc}
    \toprule[1pt]
    \multirow{2}{*}{Methods} &  \multicolumn{6}{c}{34 seen categories} & \multicolumn{6}{c}{21 unseen categories} \cr\cline{2-13} & CD-S & CD-M & CD-H & CD-Avg$\downarrow$ & DCD-Avg$\downarrow$ & F1$\uparrow$ & CD-S & CD-M & CD-H & CD-Avg$\downarrow$ & DCD-Avg$\downarrow$ & F1$\uparrow$ \cr
    \midrule[0.3pt]
    FoldingNet~\cite{yang2018foldingnet} & 1.86 & 1.81 & 3.38 & 2.35 & - & 0.139 & 2.76 & 2.74 & 5.36 & 3.62 & - & 0.095\\
    PCN~\cite{yuan2018pcn} & 1.87 & 1.81 & 2.97 & 2.22 & 0.624 & 0.150 & 3.17 & 3.08 & 5.29 & 3.85 & 0.644 & 0.101\\
    TopNet~\cite{yuan2018pcn} & 1.77 & 1.61 & 3.54 & 2.31 & - & 0.171 & 2.62 & 2.43 & 5.44 & 3.50 & - & 0.121\\
    PFNet~\cite{huang2020pf} & 3.16 & 3.19 & 7.71 & 4.68 & - & 0.347 & 5.29 & 5.87 & 13.33 & 8.16 & - & 0.322\\
    GRNet~\cite{xie2020grnet} & 1.26 & 1.39 & 2.57 & 1.74 & 0.600 & 0.251 & 1.85 & 2.25 & 4.87 & 2.99 & 0.625 & 0.216\\
    PoinTr~\cite{yu2021pointr} & 0.76 & 1.05 & 1.88 & 1.23 & 0.575 & 0.421 & 1.04 & 1.67 & 3.44 & 2.05 & 0.604 & 0.384\\
    SeedFormer~\cite{zhou2022seedformer} & 0.48 & 0.70 & 1.30 & 0.83 & 0.561 & 0.452 & \textbf{0.61} & 1.07 & 2.35 & 1.34 & 0.586 & 0.402\\
    \midrule[0.3pt]
    Ours & \textbf{0.46}  & \textbf{0.65} & \textbf{1.13}  & \textbf{0.75} & \textbf{0.538} & \textbf{0.457} &\textbf{ 0.61} & \textbf{1.05} & \textbf{2.19} & \textbf{1.28} & \textbf{0.554} & \textbf{0.427}\\
    \bottomrule[1pt]
    \end{tabular}
\end{table*}

\subsection{Dataset and Evaluation Metric}
We first use the PCN~\cite{yuan2018pcn} and ShapeNet-55/34~\cite{yu2021pointr} dataset for evaluation.
To ensure a fair comparison, we follow the same experiment settings as previous methods~\cite{yu2021pointr,zhou2022seedformer}. 
PCN contains shapes from 8 categories in ShapeNet~\cite{chang2015shapenet}. The ground-truth complete point cloud has 16,384 points and the partial input has 2,048 points. 
ShapeNet-55~\cite{yu2021pointr} is also created based on ShapeNet~\cite{chang2015shapenet} and contains shapes from 55 categories. The ground-truth point cloud has 8,192 points and the partial input has 2,048 points.
ShapeNet-34 contains 34 categories for training and leaves 21 unseen categories for testing, which is used for evaluation of generalization ability on novel categories that are unseen during training. 
Secondly, to evaluate the generalization ability in real-world scenarios, we test our method on both KITTI~\cite{geiger2013vision} and ScanNet~\cite{dai2017scannet}, which contain partial point clouds extracted from LiDAR scans and RGB-D scans, respectively. Specifically, we test on 2,401 KITTI cars extracted by \cite{yuan2018pcn}, and 100 chair point clouds from ScanNet.
We use CD, Density-aware CD (DCD)~\cite{wu2021balanced}, and F1-Score as evaluation metrics. Following the recent work~\cite{zhou2022seedformer}, we report the $\displaystyle \ell ^{1}$ version of CD for PCN and the $\displaystyle \ell ^{2}$ version of CD for Shapenet-55, for an easier comparison.

\subsection{Results on the PCN Dataset}
We compare SVDFormer with state-of-the-art methods~\cite{yuan2018pcn,xie2020grnet,wang2020cascaded,zhang2020detail,yu2021pointr,xiang2021snowflakenet,wen2022pmp,yan2022fbnet,zhou2022seedformer,zhang2022point} in Table~\ref{tab:pcnl1}. CD values are courtesy of \cite{zhou2022seedformer,yu2021pointr}, while F1-Score and DCD values are computed using their pre-trained models. The quantitative results demonstrate that SVDFormer achieves almost the best performance across all metrics. Especially, our method outperforms SeedFormer by 8.06\% in DCD.

\begin{figure}[t]
  \centering
  \includegraphics[width=0.5\textwidth]{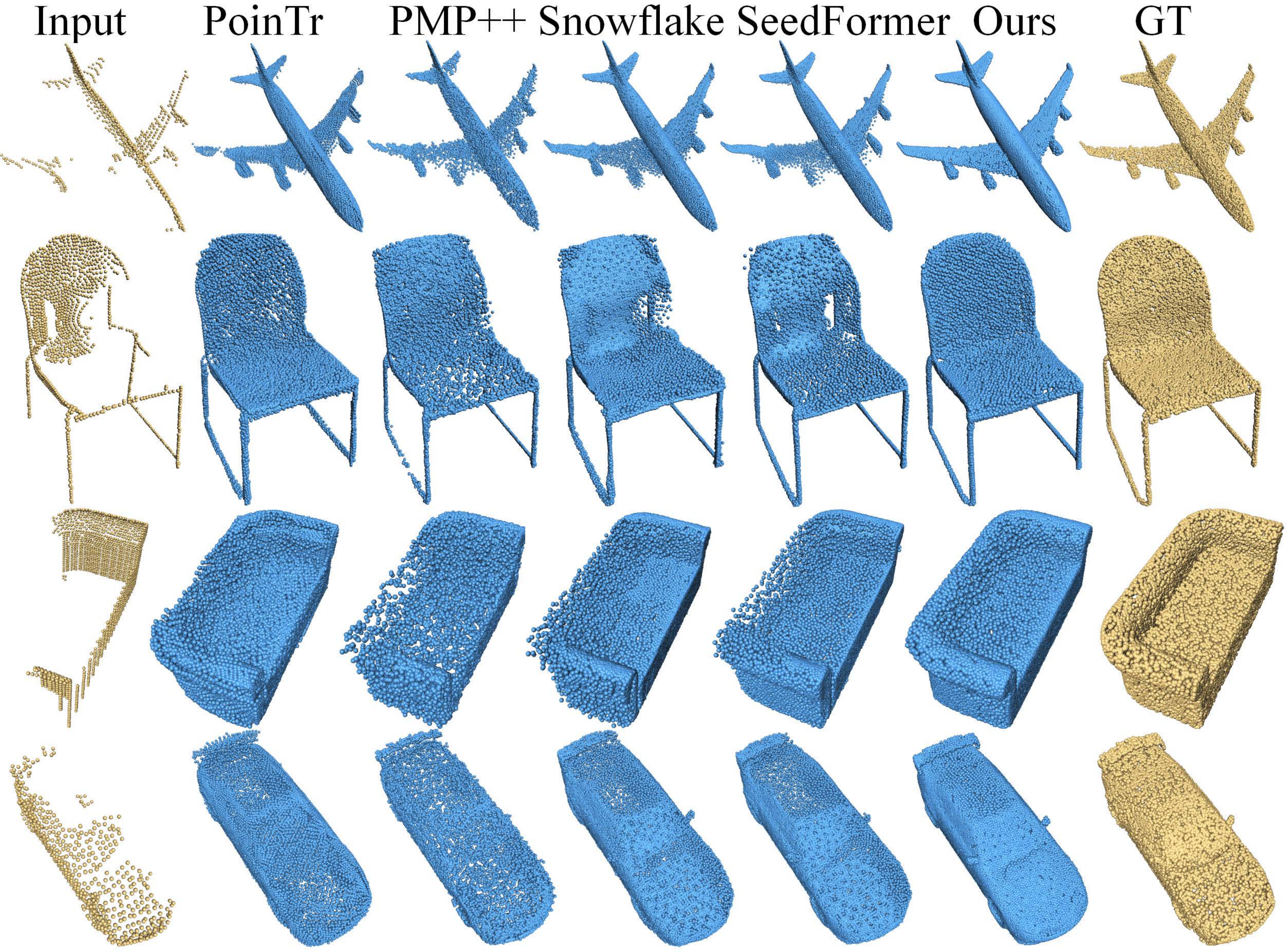}
\caption{Visual comparisons with recent methods~\cite{yu2021pointr,wen2022pmp,xiang2021snowflakenet,zhou2022seedformer} on the PCN dataset. Our method produces the most complete and detailed structures compared to its competitors.}
  \label{fig:Vis_PCN}
\end{figure}

Figure~\ref{fig:Vis_PCN} provides a visual comparison of the results produced by the different methods. In the case of the car and plane models, all methods are successful in generating the overall shapes. However, SVDFormer outperforms the other methods by producing sharper and more complete edges for detailed structures, such as plane wings and car spoilers. This is due to the generation ability of SDG. In the case of chair and couch models, SVDFormer can accurately locate missing regions and generate points among the holes in the models, leading to more faithful results.

\subsection{Results on the ShapeNet-55/34 Dataset}
\begin{figure}[t]
  \centering
  \includegraphics[width=\linewidth]{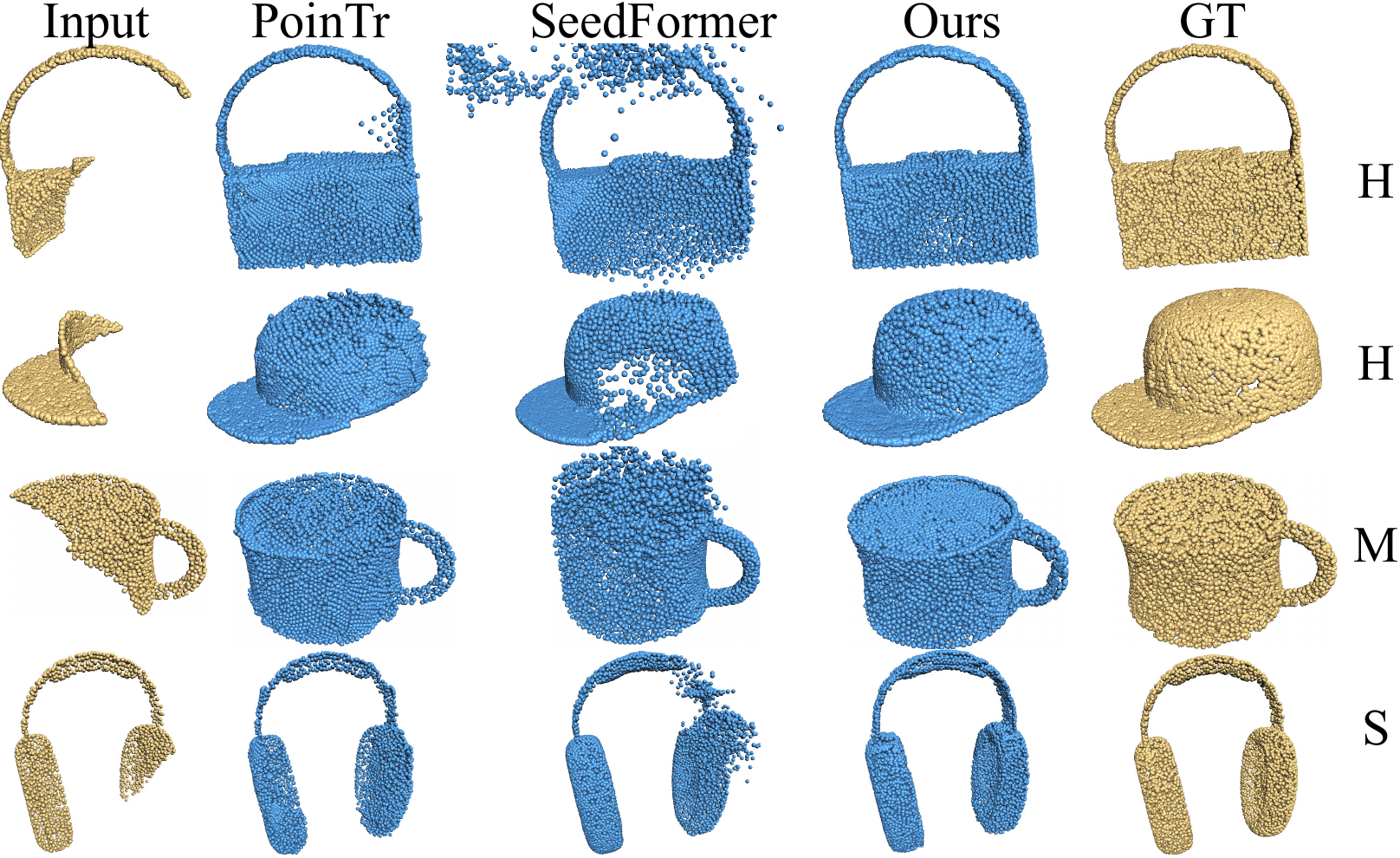}
\caption{Visual comparison with two representative approaches \cite{yu2021pointr,zhou2022seedformer} on ShapeNet-55. H (Hard), M (Moderate), and S (Simple) stand for the three difficulty levels.}
  \label{fig:vis55}
\end{figure}

The test set of ShapeNet-55 can be classified into three levels of difficulty: simple (S), moderate (M), and hard (H), which correspond to different numbers of missing points (2,048, 4,096, and 6,144).
The quantitative results are presented in Table~\ref{tab:shapenet55}, consisting of CD values for three difficulty levels and the average value of two additional metrics. Our method achieves the best result in CD across all difficulty settings. Notably, SVDFormer outperforms the state-of-the-art method SeedFormer, achieving a 12.8\% improvement in CD for the hard difficulty level. 
The results under different difficulty levels are shown in Figure~\ref{fig:vis55}. Compared to PoinTr and SeedFormer, our method produces smoother surfaces.
The visual results clearly demonstrate that SVDFormer is capable of efficiently recovering geometries from shapes with varying degrees of incompleteness.

We further evaluate SVDFormer on the ShapeNet-34 dataset. The results on both seen and unseen categories are detailed in Table~\ref{tab:shapenet34},which shows that SVDFormer achieves the best performance in terms of all three metrics.

\subsection{Results on Real-world Scans}
\begin{figure}[t]
  \centering
  \includegraphics[width=\linewidth]{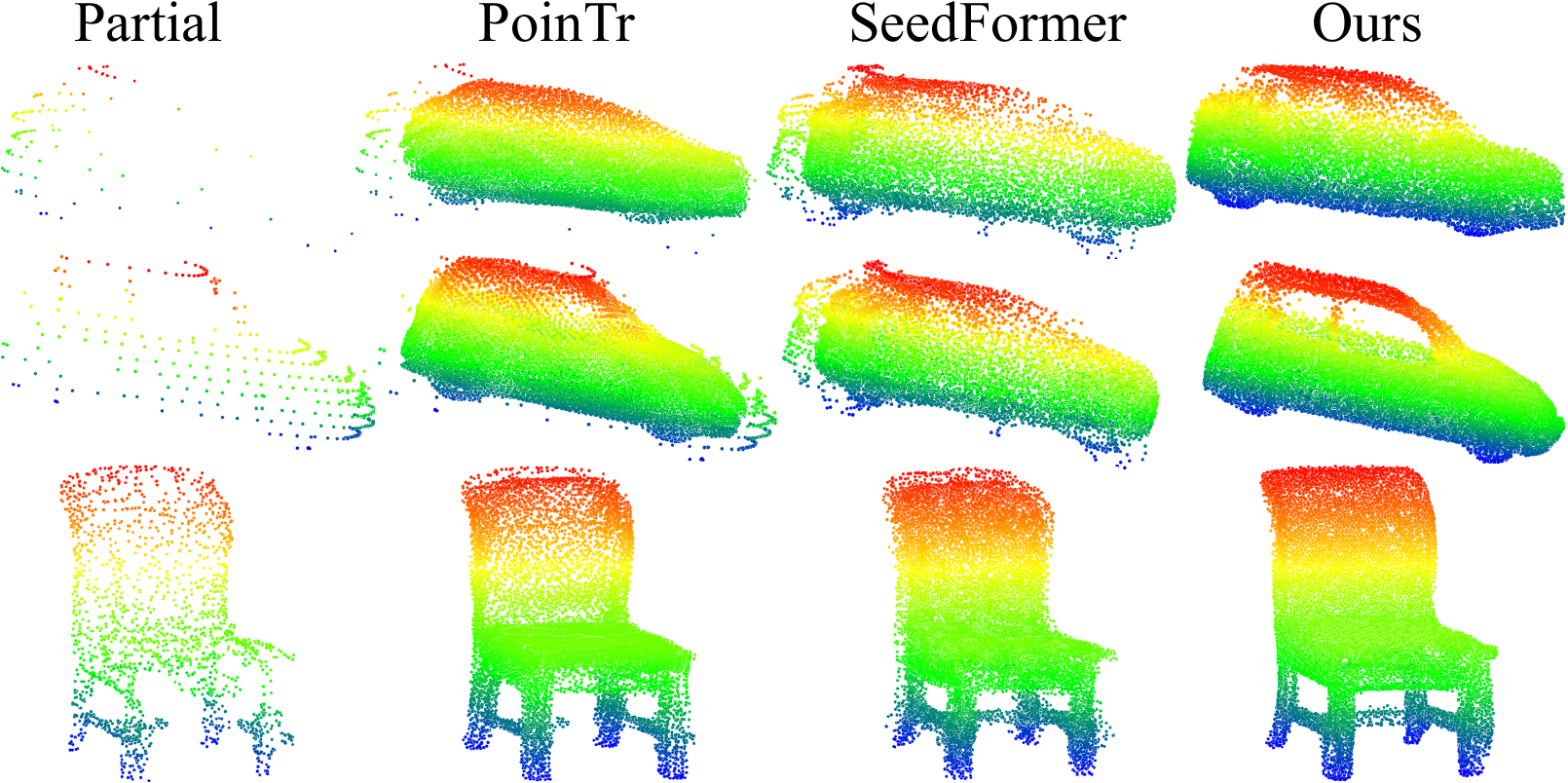}
\caption{Visual comparison on real-world scans. All results are produced by models pre-trained on PCN.}
  \label{fig:real}
\end{figure}

For the real-world scans, as there is no ground truth available for real-world partial point clouds, we evaluate the models pre-trained on PCN without fine-tuning or re-training. We report the Minimal Matching Distance (MMD)~\cite{yuan2018pcn} (see from Table~\ref{tab:realworld}) as a quantitative evaluation metric to assess the similarity of the output to a typical car/chair for the real-world scans.
Also, we demonstrate the visual comparisons in Figure~\ref{fig:real}. Our method can produce cleaner shapes with detailed structures and sharp edges. It can be concluded that our method can generate more detailed results even when the input is extremely sparse and has a different distribution from the training data.

\subsection{Ablation Studies and Discussions}
\label{ablationSec}
To ablate SVDFormer, we remove and modify the main components. All ablation variants are trained and tested on the PCN dataset. The ablation variants can be categorized as ablations on SVFNet and SDG.

\begin{table}
    \renewcommand\arraystretch{1.2}
    \footnotesize
    \centering
    \caption{Quantitative results on Real-world Scans. All the results are produced by models pre-trained on PCN (MMD $\times 10^3$).}
    \label{tab:realworld}
    \begin{tabular}{c|c|c|c|c}
    \toprule[1pt]
    Methods & GRNet~\cite{xie2020grnet} & PoinTr~\cite{yu2021pointr} & SeedFormer~\cite{zhou2022seedformer} & Ours \cr
    \midrule[0.3pt]
              KITTI~\cite{geiger2013vision}   & 5.350 & 32.854 & 1.179 & \textbf{0.967}\\
              ScanNet~\cite{dai2017scannet}   & 2.672 & 2.516 & 2.231 & \textbf{1.926} \\
              \bottomrule[1pt]
    \end{tabular}
\end{table}

\begin{table}[t]
        \renewcommand\arraystretch{1.2}
        \centering
        \caption{Effect of SVFNet. ({$\displaystyle \ell ^{1}$} CD $\times 10^3$ and F1-Score@1\%)}
        \label{tab:ablationSVFNet}
        \footnotesize
        \begin{tabular}{c|c c c}
        \toprule[1pt]
        Methods  & CD$\downarrow$ & DCD$\downarrow$ &F1$\uparrow$  \\
        \midrule[0.3pt]
        A : w/o Projection  & 6.63 & 0.547 & 0.831\\
        B : w/o Feature Fusion & 6.68 & 0.551 & 0.827\\
        Ours  & \textbf{6.54} & \textbf{0.536}& \textbf{0.841}  \\
        \bottomrule[1pt]
        \end{tabular}
\end{table}

\begin{table}[t]
        \renewcommand\arraystretch{1.2}
        \centering
        \caption{Effect of SDG. ({$\displaystyle \ell ^{1}$} CD $\times 10^3$ and F1-Score@1\%)}
        \label{tab:ablationIDTr}
        \footnotesize
        \begin{tabular}{c|c c c|c c c}
        \toprule[1pt]
        Methods & \makebox[0.12\linewidth][c]{Analysis} & \makebox[0.12\linewidth][c]{Alignment} & \makebox[0.12\linewidth][c]{Embedding} & \makebox[0.01\linewidth][c]{CD$\downarrow$} & \makebox[0.01\linewidth][c]{DCD$\downarrow$} & \makebox[0.01\linewidth][c]{F1$\uparrow$}  \\
        \midrule[0.3pt]
        C & \checkmark & \checkmark & & 6.69 & 0.549 & 0.829 \\
        D & \checkmark & & \checkmark & 6.76 & 0.552 & 0.825\\
        E & &\checkmark &  &  6.78 & 0.556  & 0.823\\
        F &\checkmark& & & 6.88 & 0.561 & 0.819\\
        Ours & \checkmark & \checkmark & \checkmark & \textbf{6.54} & \textbf{0.536} &\textbf{0.841}  \\
        \midrule[0.3pt]
        \multicolumn{4}{c|}{SeedFormer~\cite{zhou2022seedformer}} &  6.74 & 0.583 & 0.818 \\
        \multicolumn{4}{c|}{SnowflakeNet-baseline~\cite{xiang2021snowflakenet}} & 7.21 & 0.585 & 0.801\\
        \multicolumn{4}{c|}{SnowflakeNet + SDG} &  6.73 & 0.553 & 0.828 \\
        \bottomrule[1pt]
        \end{tabular}
\end{table}
\begin{figure}[t]
  \centering
  \includegraphics[width=0.95\linewidth]{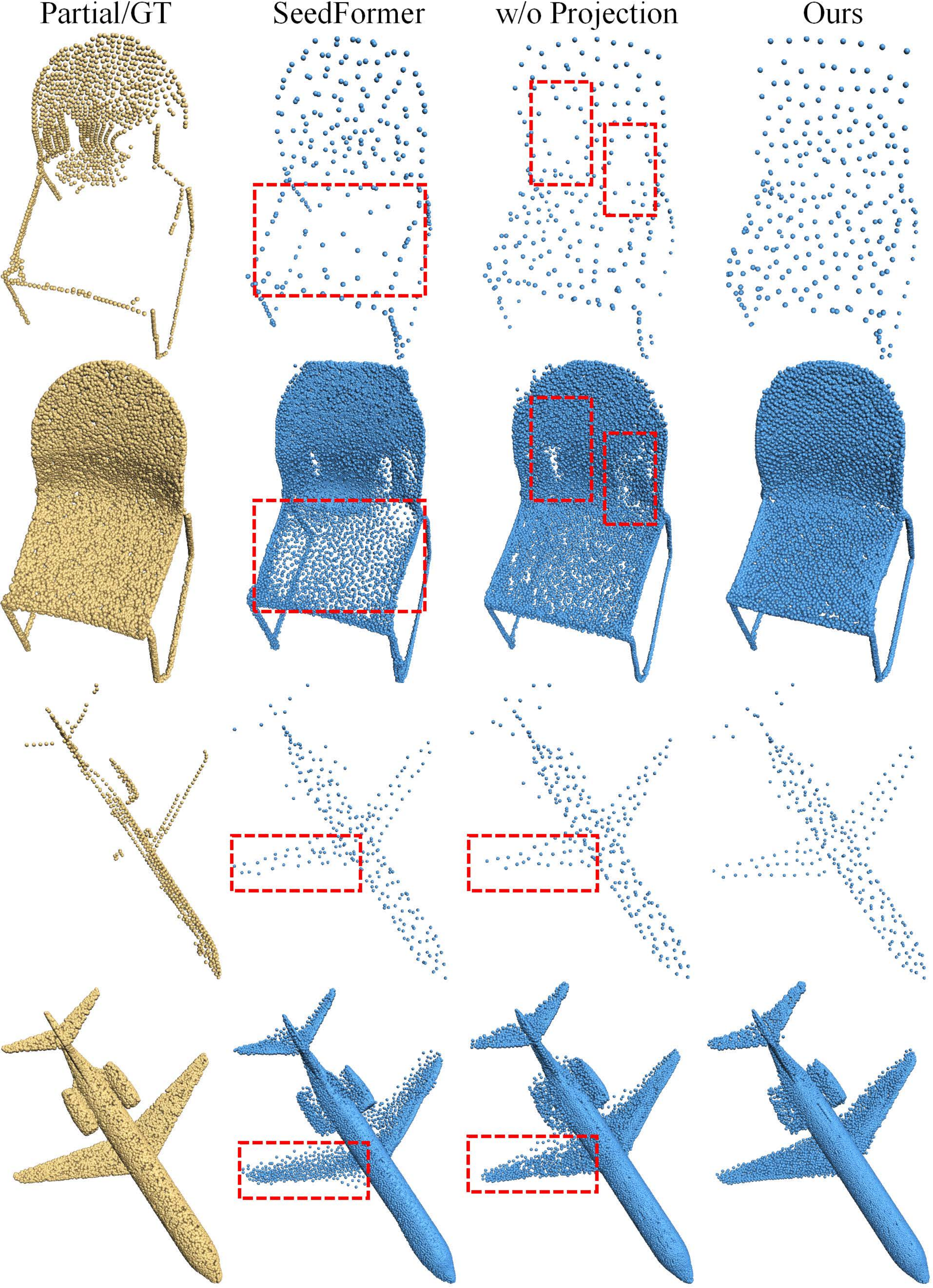}
\caption{Visual comparison of the representative coarse-to-fine method \cite{zhou2022seedformer} and our variant A (w/o projection) on two partial models. The upper results are the generated coarse point clouds.}
  \label{fig:MVFVis}
\end{figure}
\begin{figure}[h]
  \centering
  \includegraphics[width=\linewidth]{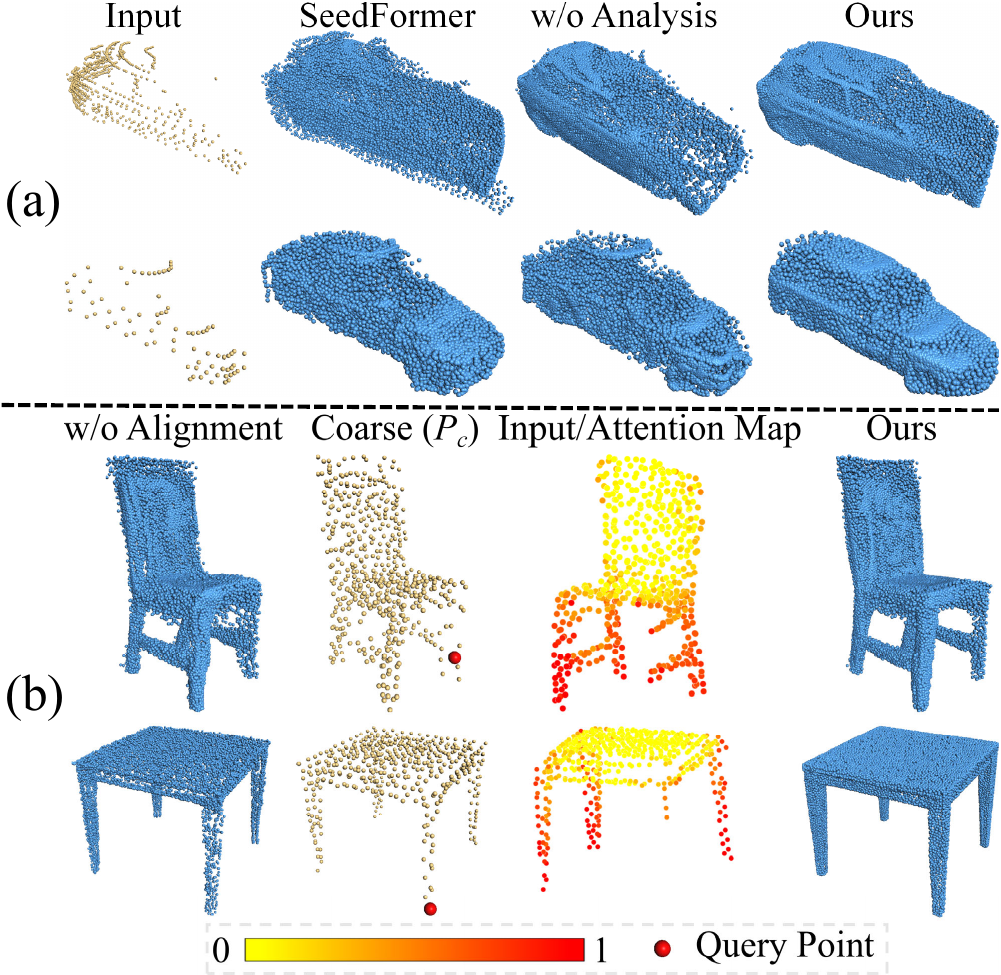}
\caption{(a): Visual comparison of~\cite{zhou2022seedformer} and variant E (w/o Structure Analysis) on LiDAR scans. (b): Visual comparison of variant E (w/o Similarity Alignment) and generated coarse point cloud $P_c$ on RGB-D scans. We select a query point (marked with red) in $P_c$ and visualize the attention map in the cross-attention layer. The redder the color, the higher the similarity.}
  \label{fig:IDTrAblation}
\end{figure}

\noindent\textbf{\textbf{Ablation on SVFNet}.}
To investigate the impact of shape descriptor extraction methods, we compare two variants of SVFNet, and the results are presented in Table~\ref{tab:ablationSVFNet}. In the variant A, we remove the input depth maps, and the completion performance is limited by relying only on 3D coordinates to understand shapes. In the variant B, we evaluate the importance of our Feature Fusion module by replacing the fusion of different inputs with late fusion, which directly concatenates $F_p$ and $F_v$. We observe an evident drop in performance, indicating that the proposed SVFNet can effectively fuse cross-modal features. 

Furthermore, to conduct a more thorough analysis of the effectiveness of our SVFNet, we generate visualizations of the results produced by our approach, our variant A, and SeedFormer, which also employ the coarse-to-fine paradigm. In Figure~\ref{fig:MVFVis}, we present the results alongside the coarse point cloud generated directly by SVFNet (patch seeds of \cite{zhou2022seedformer}). Our analysis reveals that during the initial coarse completion stage, both SeedFormer and the variant A produce suboptimal results, such as generating too few points in missing areas. This presents a challenge for the refinement stage, making it difficult to produce satisfactory final results. Our SVFNet overcomes this challenge by leveraging multiple viewpoints to observe partial shapes. By doing so, our method is able to locate missing areas and generate a compact global shape, leading to the production of fine details in the final results. See the supplementary for additional ablation experiments for the number of views.

\noindent\textbf{Ablation on SDG.}
Table~\ref{tab:ablationIDTr} compares different variants of SVDFormer on the SDG module. In the variant C, we remove the Incompleteness Embedding of SDG, which results in a higher CD value and a lower F1-Score, indicating that the ability to perceive the incompleteness degree of each part is crucial for the model's performance. In the variants D and E, we completely remove the Similarity Alignment and Structure Analysis path from SDG, respectively. The results show that the performance of the model drops when any one of these paths is removed.

To better understand and analyze SDG, we show more visual results in Figure~\ref{fig:IDTrAblation}. 
Specifically, we investigate the effectiveness of the Structure Analysis path and the Similarity Alignment unit by comparing the performance of different variants of the model on real-world scans. 
In Figure~\ref{fig:IDTrAblation} (a), our method can generate plausible shapes, while the variant E and SeedFormer produce undesired structures due to over-preserving the partial input. 
This result proves the importance of the Structure Analysis path, particularly when the input contains limited information. In Figure~\ref{fig:IDTrAblation} (b), we compare the results of our method with the variant D. We show the generated coarse shape and select a query point (missed in input) in $P_c$. We then visualize the attention map in the cross-attention layer to demonstrate the effectiveness of the Similarity Alignment unit. The results show that, for shapes with highly-similar regions, the Similarity Alignment unit can locate similar geometries of short or long-range distances, leading to finer details.

\noindent\textbf{Extending SDG to other methods.} 
In addition, we evaluate the generation ability of SDG by replacing the SPD in SnowflakeNet~\cite{xiang2021snowflakenet} with SDG. As presented in Table~\ref{tab:ablationIDTr}, SnowflakeNet achieves better performance in terms of all metrics, when paired with our SDG module. This indicates that our disentangled refiner has better generation ability.

\noindent\textbf{Complexity Analysis. } 
We show the complexity analysis in Table~\ref{tab:complexity}, where the inference time on a single NVIDIA 3090 GPU, number of parameters, and the results on ShapeNet-55 are shown. The comparison indicates that our method achieves a trade-off between cost and performance.

\begin{table}[t]
        \renewcommand\arraystretch{1.2}
        \centering
        \caption{Complexity analysis. We compare the inference time (ms) and the number of parameters (Params) of our method and three classical methods on ShapeNet-55. Our method achieves a balance between computation cost and performance.}
        \label{tab:complexity}
        \small
        \begin{tabular}{c|c c| cc} 
        \toprule[1pt]
        Methods  & Time & Params & CD$\downarrow$ & DCD$\downarrow$  \\
        \midrule[0.3pt]
        GRNet~\cite{xie2020grnet}  & 10.67ms & 73.15M & 1.97 & 0.592 \\
        PoinTr~\cite{yu2021pointr}  & 12.36ms & 30.09M & 1.07 & 0.575 \\
        SeedFormer~\cite{zhou2022seedformer}  & 40.63ms & 3.24M & 0.92 & 0.558 \\
        Ours  & 23.11ms &  19.62M & 0.83 & 0.546 \\
        \bottomrule[1pt]
        \end{tabular}
\end{table}

\label{secEXP}

\section{Conclusion}
We propose SVDFormer for point cloud completion. We start by identifying the main challenges in the completion and developing new solutions for each of them. SVDFormer leverages self-projected multi-view analysis to comprehend the overall shape and effectively perceive missing regions. Furthermore, we introduce a decoder called Self-structure Dual-generator that breaks down the shape refinement process into two sub-goals, resulting in a disentangled but improved generation ability. Experiments on various shape types demonstrate that SVDFormer achieves the state-of-the-art performance on point cloud completion.
\label{secCon}

\vspace{-0.1cm}
\section*{Acknowledgements}
The work described in this paper was supported by the National Natural Science Foundation of China (No. 62172218), the Research Grants Council of the Hong Kong Special Administrative Region, China (No. UGC/FDS16/E14/21), the Shenzhen Science and Technology Program (No. JCYJ20220818103401003, No. JCYJ20220530172403007), the Natural Science Foundation of Guangdong Province (No. 2022A1515010170), and Hong Kong RGC General Research Fund (No. 15218521).

{\small
\bibliographystyle{ieee_fullname}
\bibliography{egbib}
}
\newpage

\renewcommand\thesection{\Alph{section}}
In this supplementary material, we provide more detailed information to complement the main manuscript. 
Specifically, we first introduce the implementation details, including network architecture details and experimental settings. 
Then, we conduct more ablation studies to analyze our method.
Next, we provide some failure cases and a discussion on the limitations of our work.
Finally, we present additional quantitative and qualitative results. 
\setcounter{section}{0}
\section{Detailed Settings}
\noindent\textbf{\textbf{Network implementation details}.}
We apply perspective projection to get the depth maps with the resolution of 224 × 224 from three orthogonal views. We directly feed the projected depth maps to the network without applying any color mapping enhancement.
In SVFNet, we use PointNet++~\cite{qi2017pointnet++} to extract features from point clouds. The detailed architecture is: 
$\emph{SA}(C = [3,64,128], N = 512, K = 16)\rightarrow\emph{SA}(C = [128,256], N = 128, K = 16)\rightarrow\emph{SA}(C = [512,256])$. The final feature dimension of ResNet18~\cite{he2016deep} is set to 256. The dimension of the embed query, key, and value in View Augment is set to 256. After concatenation, we get the shape descriptor $F_g$ with 512 channels. 
We use a self-attention layer of 512 hidden feature dimensions followed by an MLP to regress the coarse points $P_C$. The merged point cloud $P_0$ has 512 and 1024 points for PCN and ShapeNet-55, respectively.

During refinement, we set the upsampling rates \{$r_1$, $r_2$\} of the two SDGs as \{4,8\} and \{2,4\} for PCN and ShapeNet-55, respectively.
We adopt EdgeConv~\cite{wang2019dynamic} to extract local features from $P_{in}$. The detailed architecture is:
$\emph{EdgeConv}(C = [3,64], K = 16)\rightarrow\emph{FPS}(2048,512)\rightarrow\emph{EdgeConv}(C = [64,256], K = 8)$. We use a shared-weights architecture above in the two SDGs. After obtaining $F_Q$ and $F_H$, we use a decoder composed of two self-attention layers (one in the ShapeNet-55 experiments) to further analyze the coarse shapes. The hidden feature dimensions of self-attention layers are set as [768, 128$r_1$] and [512, 128$r_2$] in the two SDGs, thus producing $F_{l}\subseteq\mathbb{R}^{N\times 256r}$. $F_{l}$ is then passed to an MLP and reshaped to ${rN\times 128}$. Finally, the coordinates offset is predicted by an MLP with feature dimensions of [128, 64, 3].

\begin{figure}[t]
  \centering
  \includegraphics[width=0.8\linewidth]{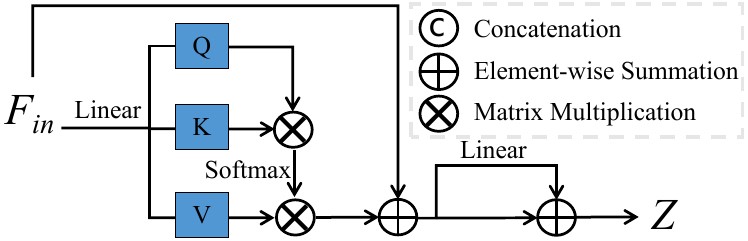}
\caption{The calculation process of Self-Attention.}
  \label{fig:attn}
\end{figure}

\noindent\textbf{\textbf{Usage of attention}.}
In our method, the self-attention layer is used to generate $P_c$ in SVFNet and decode $F_Q$ and $F_H$ in SDG. We also use a cross-attention layer to find the geometric similarity.
In our experiments, we implement the self-attention module and the cross-attention module following the same transformer architecture~\cite{vaswani2017attention}. The point-wise features are regarded as sequence data. The calculation procedure is illustrated in Figure~\ref{fig:attn}. Given the input feature $F_{in}=\{f_i\}^{N_{l-1}}_{i=1}$, the output feature matrix $Z=\{z_i\}^{N_{l-1}}_{i=1}$ is calculated as :
\begin{equation} \label{eqn1}
  \begin{split}
   z_i &= h_i + Linear(h_i)\\
   h_i &= b_i + f_i\\
   b_i &= \sum_{j=1}^{N_{l-1}}{a_{i,j}(f_jW_V)}\\
   a_{i,j} &= Softmax((f_iW_Q)(f_jW_K)^T)
  \end{split},
\end{equation}

\begin{figure*}[h]
  \centering
  \includegraphics[width=\textwidth]{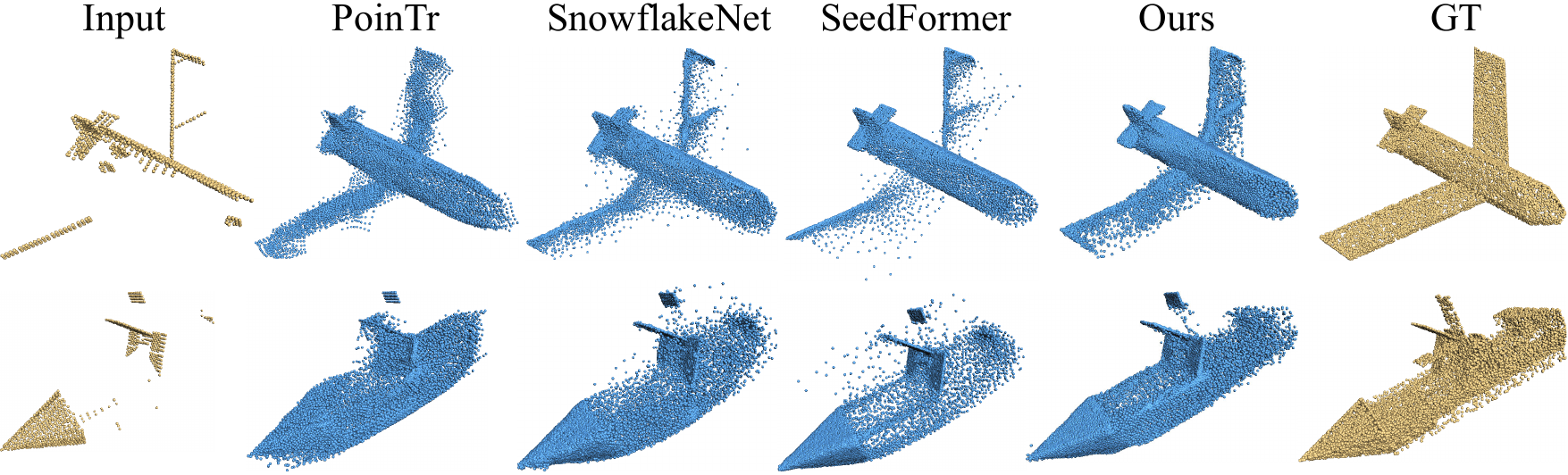}
\caption{Example of failure cases.}
  \label{fig:failurecases}
\end{figure*}
\begin{table}
        \renewcommand\arraystretch{1.1}
        \footnotesize
        \centering
        \caption{Results and inference time of more ablation variants on PCN. ({$\displaystyle \ell ^{2}$} CD $\times 10^3$ and F1-Score@1\%)}
        \label{tab:ablationNumber}
        \begin{tabular}{c|c c c|c}
        \toprule[1pt]
        Variants  & CD$\downarrow$ & DCD$\downarrow$ &F1$\uparrow$ & Time  \\
        \midrule[0.3pt]
        1 View  & 6.58 & 0.538 & 0.835 & 24.86ms\\
        3 Views (Ours) & 6.54 & 0.536 & \textbf{0.841} & 26.55ms\\
        6 Views & 6.55 & 0.536 & 0.840 & 27.11ms\\
        \midrule[0.3pt]
        Random Projection (inference) & 6.58 & 0.537 & 0.838 & 26.25ms\\
        \midrule[0.3pt]
        Encoder in SpareNet~\cite{xie2021style} & 6.66 & 0.551  & 0.825 & 37.26ms\\
        ResNet-50~\cite{he2016deep} as the 2D backbone & \textbf{6.52}  & \textbf{0.535}  &  \textbf{0.841} & 31.37ms\\
        Vit-B/16~\cite{dosovitskiy2021an} as the 2D backbone & 6.56 & 0.543 & 0.837 & 34.16ms\\
        \bottomrule[1pt]
        \end{tabular}
\end{table}

\noindent\textbf{\textbf{Experiment and training settings}.}
The network is implemented using PyTorch~\cite{paszke2019pytorch} and trained with the Adam optimizer~\cite{kingma2014adam} on NVIDIA 3090 GPUs. 

For training on the PCN dataset~\cite{yuan2018pcn}, the initial learning rate is set to 0.0001 and decayed by 0.7 for every 40 epochs. The batch size is set to 12. It takes 400 epochs for convergence. 
Since the point coordinates in PCN are normalized to [-0.5, 0.5], the depth maps are projected at a distance of 0.7 in order to observe the whole shape. 
To ensure that the input point cloud contains exactly 2048 points, we take a subset for point clouds with more than 2048 points and randomly duplicate points for those with less than 2048 points.

For training on ShapeNet-55/34~\cite{yu2021pointr}, the number of missing points is randomly selected from 2048 to 6192.
The initial learning rate is set to 0.0001 and decayed by 0.98 for every 2 epochs. The batch size is set to 16. It takes 300 epochs for convergence. 
The point coordinates in ShapeNet-55 are normalized to [-1.0, 1.0]. Therefore, $D$ are projected at a distance of 1.5. 
Following \cite{xiang2021snowflakenet,zhou2022seedformer}, We use a partial matching strategy, which includes setting a larger resolution for $P_{0}$ and adding a partial matching loss~\cite{wen2021cycle4completion}.

\section{Ablation Studies}

\begin{table*}[htb]
    \renewcommand\arraystretch{1.0}
    \centering
    \caption{DCD results on the PCN dataset. Lower is better.}
    \label{tab:pcnDCD}
    \small
    \begin{tabular}{c|cccccccc|c}
    \toprule[1pt]
    Methods & Plane & Cabinet & Car & Chair & Lamp & Couch & Table & Boat & Avg \cr
    \midrule[0.3pt]
              GRNet~\cite{xie2020grnet}  & 0.688 & 0.582 & 0.610 & 0.607 & 0.644 & 0.622 & 0.578 & 0.642 & 0.622\\
              PoinTr~\cite{yu2021pointr}  & 0.574 & 0.611 & 0.630 & 0.603 & 0.628 & 0.669 & 0.556 & 0.614 & 0.611\\
              SnowflakeNet~\cite{xiang2021snowflakenet} & 0.560 & 0.597 & 0.603 & 0.582 & 0.598 & 0.633 & 0.521 & 0.583 & 0.585 \\
              PMP-Net++~\cite{wen2022pmp} & 0.600 & 0.605 & 0.614 & 0.613 & 0.610 & 0.647 & 0.577 & 0.622 & 0.611\\
              Seedformer~\cite{zhou2022seedformer}  & 0.557 & 0.592 & 0.598 & 0.579 & 0.585 & 0.626 & 0.520 & 0.605 & 0.583\\
              \midrule[0.3pt]
              Ours & \textbf{0.506} & \textbf{0.549} & \textbf{0.559} & \textbf{0.524} & \textbf{0.535} & \textbf{0.579} & \textbf{0.472} & \textbf{0.562}& \textbf{0.536} \\
              \bottomrule[1pt]
    \end{tabular}
\end{table*}

\begin{table*}[h!t]
    \renewcommand\arraystretch{1.0}
    \centering
    \caption{F1-Score@1\% on the PCN dataset. Higher is better.}
    \label{tab:pcnf1}
    \small
    \begin{tabular}{c|cccccccc|c}
    \toprule[1pt]
    Methods & Plane & Cabinet & Car & Chair & Lamp & Couch & Table & Boat & Avg \cr
    \midrule[0.3pt]
              GRNet~\cite{xie2020grnet}  & 0.843 & 0.618 & 0.682 & 0.673 & 0.761 & 0.605 & 0.751 & 0.750 & 0.708\\
              PoinTr~\cite{yu2021pointr}  & 0.915 & 0.665 & 0.718 & 0.710 & 0.798 & 0.632 & 0.796 & 0.797 & 0.754\\
              SnowflakeNet~\cite{xiang2021snowflakenet} & 0.941 & 0.695 & 0.745 & 0.776 & 0.858 & 0.691 & 0.867 & 0.834 & 0.801 \\
              PMP-Net++~\cite{wen2022pmp} & 0.941 & 0.660 & 0.721 & 0.754 & 0.860 & 0.657 & 0.822 & 0.830 & 0.781\\
              Seedformer~\cite{zhou2022seedformer}  & 0.950 & 0.700 & 0.753 & 0.803 & 0.885 & 0.712 & 0.884 & 0.850 & 0.818\\
              \midrule[0.3pt]
              Ours & \textbf{0.962} & \textbf{0.738} & \textbf{0.792} & \textbf{0.833} & \textbf{0.897} & \textbf{0.746} & \textbf{0.901} & \textbf{0.863}& \textbf{0.841} \\
              \bottomrule[1pt]
    \end{tabular}
\end{table*}

\begin{figure*}[hp]
  \centering
  \includegraphics[width=\textwidth]{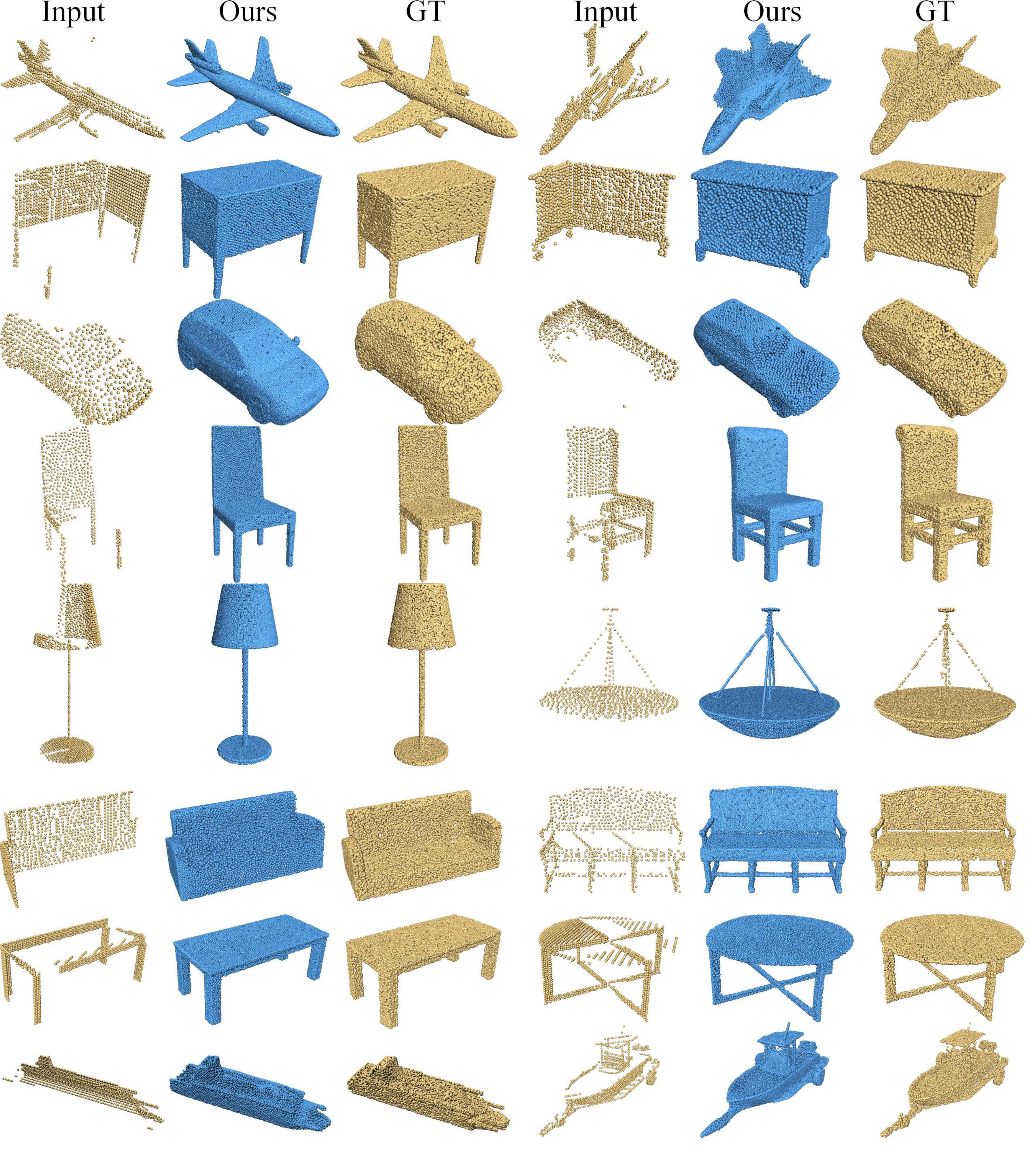}
\caption{Visual results on the PCN dataset.}
  \label{fig:pcnbest}
\end{figure*}
\begin{figure*}[hp]
  \centering
  \includegraphics[width=\textwidth]{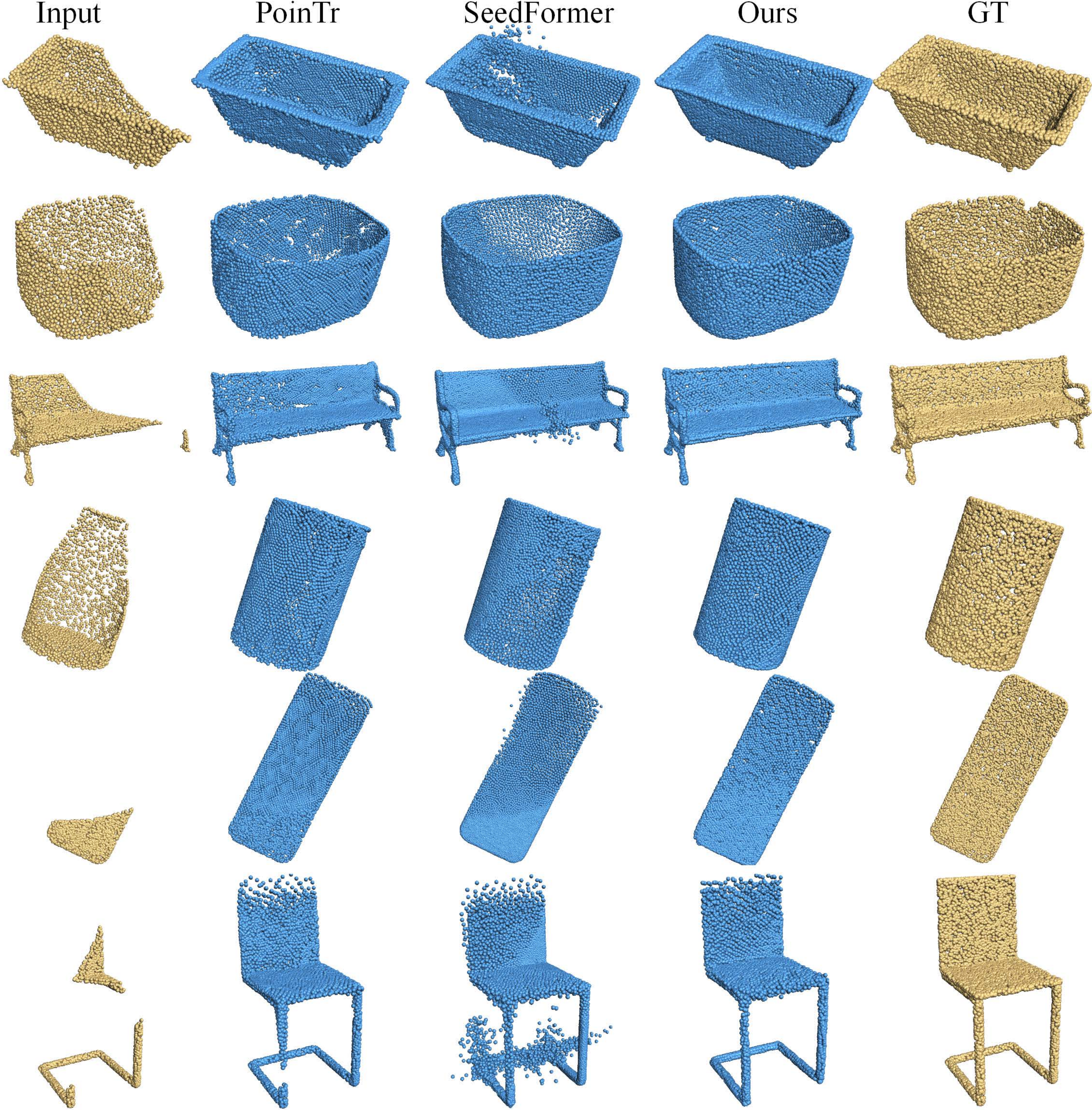}
\caption{Visual Comparison with two representative approaches \cite{yu2021pointr,zhou2022seedformer} on ShapeNet-55.}
  \label{fig:55compare}
\end{figure*}

\begin{figure*}[hp]
  \centering
  \includegraphics[width=\textwidth]{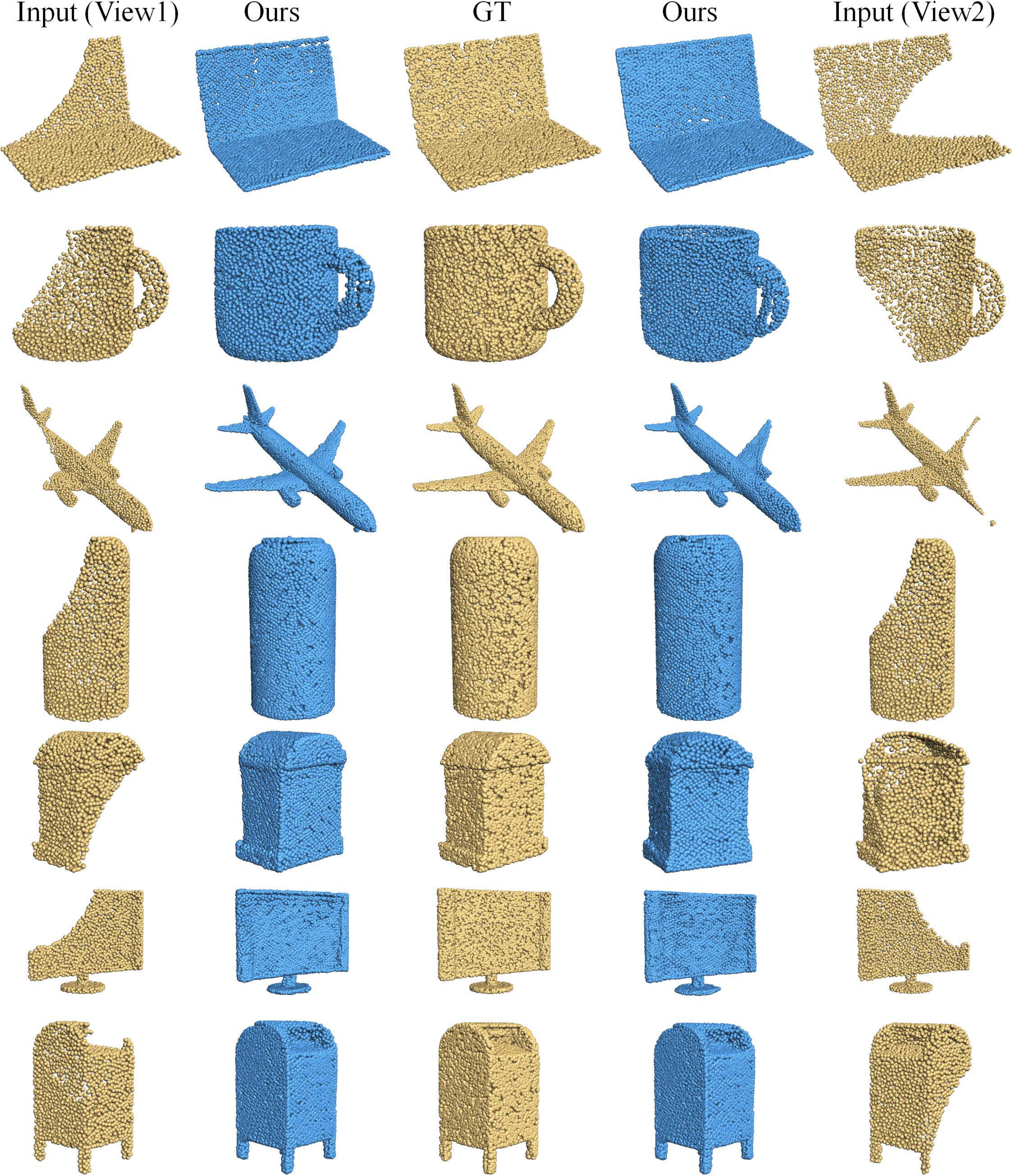}
\caption{More visual results on ShapeNet-55. We show results when the partial input are generated from two viewpoints.}
  \label{fig:55more}
\end{figure*}

\noindent\textbf{\textbf{Ablation on the number of projections}.}
We conduct an ablation experiment on the number of depth maps $D$ in SVFNet. The depth maps $D$ are projected from 1, 3, and 6 orthogonal views, respectively. The results on PCN are shown in Table~\ref{tab:ablationNumber}. To balance the trade-off between effectiveness and computational consumption, we conduct all experiments using three views. This choice allows us to capture sufficient information from the point clouds while keeping the computational cost manageable. 

\noindent\textbf{\textbf{Ablation on choice of coordinate systems}.}
To testify the robustness of our method, during inference, we introduced random variations to the projection, including camera view angle offsets ranging from 0 to 10 degrees and observation distance displacements ranging from 0 to 0.1. The result reported in the 5th row of Table~\ref{tab:ablationNumber} shows that the performance will not significantly drop with random projections. 

\noindent\textbf{\textbf{Ablation of different encoders}.}
We testify the design of encoder to further demonstrate the effect of our self-view fusion feature extractor. We first replace the SVFNet with the encoder in SpareNet~\cite{xie2021style}, which contains layers of channel-attentive EdgeConv, to re-produce the shape descriptor $F_g$. We report the new results in Table~\ref{tab:ablationNumber}, which demonstrates that our self-view fusion feature extractor achieves better performance than existing encoder while having a tolerable computation cost. 
In addition, we ablate the choice of 2D backbone in the SVFNet. To be specific, We replace it with ResNet-50~\cite{he2016deep} and the vision transformer (ViT-B/16)~\cite{dosovitskiy2021an}, respectively. We find that a larger CNN-based 2D backbone can slightly improve the performance while introducing more computation cost. Moreover, using the ViT results in unsatisfactory performance. This could be attributed to the fact that the entire model was trained from scratch, and larger models may not perform optimally with a limited amount of 3D training data.


\section{Failure Cases and Limitations}
Figure~\ref{fig:failurecases} displays the failure cases we observed.
It's worth noting that in cases where input shapes lack irregular structures that are uncommon in the training dataset (such as the water wheel of a watercraft), the network may not be capable of producing satisfactory results.
Nevertheless, our method still outperforms state-of-the-art (SOTA) methods~\cite{yu2021pointr,xiang2021snowflakenet,zhou2022seedformer} when dealing with simple geometric structures, like the body of the watercraft.
Our SDG incorporates a Structure Analysis unit that leverages learned priors to complete shapes. However, its effectiveness may be constrained by the limited amount of available training data. Given that transformers have demonstrated effectiveness in scenarios with abundant training data, pretraining with large-scale 2/3D datasets could be a promising approach to address this limitation.

\section{Additional Results}
More detailed quantitative results for individual cases are available in Tables~\ref{tab:pcnDCD} and \ref{tab:pcnf1}. Our method achieves the best DCD and F1-score on each category of the PCN dataset. In addition, we show more visual results in Figures~\ref{fig:pcnbest}, \ref{fig:55compare}, and \ref{fig:55more}. In Figure~\ref{fig:pcnbest}, we present two partial point clouds on each category of the PCN dataset. In Figure~\ref{fig:55compare}, we present six partial point clouds of ShapeNet-55 and compare the results with two representative approaches~\cite{yu2021pointr,zhou2022seedformer}. Our method generates more compact overall shapes and richer details. Also, we visualize more results in Figure~\ref{fig:55more}, where the partial shapes are generated from two different viewpoints.


\ificcvfinal\thispagestyle{empty}\fi

\end{document}